\documentclass[english]{ieej-e-samcon}

\usepackage{amsmath}
\usepackage{graphicx}
\usepackage[varg]{txfonts}

\usepackage{cite}

\usepackage{algorithmic}
\usepackage{graphicx}
\usepackage{textcomp}
\usepackage{xcolor}

\usepackage{multirow}
\usepackage{threeparttable}
\usepackage{bm}
\usepackage{svg}

\usepackage[normalem]{ulem}
\usepackage{cite}
\usepackage{makecell}
\usepackage[subrefformat=parens]{subcaption}

\YEAR{2025}
\title{Decoupled Scaling 4ch Bilateral Control on the Cartesian coordinate \\ by 6-DoF Manipulator using Rotation Matrix}
\authorlist{%
 \authorentry[yamane.koki.td@alumni.tsukuba.ac.jp]{Koki Yamane}{n}{tsukuba}
 \authorentry{Sho Sakaino}{S}{tsukuba}
 \authorentry{Toshiaki Tsuji}{S}{saitama}
}
\affiliate[tsukuba]
 {University of Tsukuba, \\
  1--1--1, Tennodai, Tsukuba, Ibaraki, Japan 305--8577}
\affiliate[saitama]
 {Saitama University, \\
  Shimo-Okubo 255, Sakura-ku, Saitama-sh, Saitama, Japan 338--8570}

\begin{document}
\begin{abstract}
Four-channel bilateral control is a method for achieving remote control with force feedback and adjustment operability by synchronizing the positions and forces of two manipulators. This is expected to significantly improve the operability of the remote control in contact-rich tasks.
Among these, 4-channel bilateral control on the Cartesian coordinate system is advantageous owing to its suitability for manipulators with different structures and because it allows the dynamics in the Cartesian coordinate system to be adjusted by adjusting the control parameters, thus achieving intuitive operability for humans.
This paper proposes a 4-channel bilateral control method that achieves the desired dynamics by decoupling each dimension in the Cartesian coordinate system regardless of the scaling factor.
\end{abstract}
\maketitle

\section{Introduction}
Four-channel bilateral control is a method for achieving remote control with force feedback and adjustment operability by synchronizing the positions and forces of two manipulators. It is expected to significantly improve the operability of remote control for contact-rich tasks, and in recent years, it has also been used as a data collection method in imitation learning.
Among these, the 4-channel bilateral control on the Cartesian coordinate system is advantageous in that it can be used for manipulators with different structures and that the dynamics in the Cartesian coordinate system can be adjusted by adjusting the control parameters, thus achieving intuitive operability for humans.

However, achieving high operability by controlling a Cartesian coordinate system remains challenging. In the case of joint space control, all complex interactions between joints are treated as unknown disturbances, and a certain degree of control can be achieved by combining a linear control system with a classical single-input single-output (SISO) system. However, when designing a control system in the Cartesian coordinate system, the position and posture of the manipulator's end-effector are expressed in a three-dimensional special Euclidean group (SE(3)), which has different properties from the vector spaces commonly used in traditional control methods, such as noncommutativity and the fact that addition is not defined. Therefore, it is not possible to use classical control design methods that assume vector spaces as they are. It is possible to approximate the vector space and perform control based on the assumption that the position error remains within a small interval close to zero. However, the error becomes particularly large when scaling bilateral control, which synchronizes the position and force by multiplying each by a constant. This paper proposes a 4-channel bilateral Cartesian coordinate scaling control method that achieves the desired dynamics by decoupling each dimension in the Cartesian coordinate system regardless of the scaling factor.

\section{Related Works}
Research on the controller design for bilateral control systems has been conducted for many years.
Tsuji~\textit{et al.} proposed a controller design method for a bilateral control system based on ``functionality''~\cite{tsuji2006controller}, which designs a controller on the coordinate that is transformed to the space suited to the control objective.
Sakaino~\textit{et al.} proposed an oblique coordinate control method that achieved decoupled scaling of 4-channel bilateral control for multi-degree-of-freedom (DOF) manipulators~\cite{sakaino2011multi}.
However, these experiments were conducted on 1 or 2 DoF hardware, the controller was not operated in the Cartesian coordinate system, and the method did not consider issues such as the noncommutativity of 3D rotations.

In recent years, bilateral control using multi-DOF manipulators with six or more degrees of freedom has been actively studied.
Suzuki~\textit{et al.} developed a surgical robot controlled by a 4-channel bilateral control~\cite{suzuki2018development} using fiber Bragg grating based force sensors, and Yilmaz~\textit{et al.} reported the results of implementing 4-channel bilateral control with Cartesian coordinate scaling in the da Vinci research kit ~\cite{yilmaz2023sensorless}.
However, these studies did not rigorously address the noncommutativity of 3D rotation and the interference of the dynamics of the two robots.

Michel~\textit{et al.}~\cite{michel2024passivity} proposed a bilateral control method that rigorously handles 3D rotation using a representation based on Lie algebra.
However, this study used force-feedback bilateral control instead of 4-channel bilateral control.
This teleoperation method used force control for the leader robot and position control for the follower robot.
Therefore, this study does not consider the complex interference between position control and force control.

This study proposed a control method that compensates for interfering elements and sets the error dynamics in each dimension independently while addressing the non-commutativity of 3D rotation.


\section{Preliminary}

\subsection{Rotation Matrix and Rotation Vector}
Since 3x3 orthogonal matrices with a determinant of one can represent 3D rotation, this is called a ``rotation matrix.''
The set of rotation matrices SO(3) is defined as
\begin{align}
SO(3) \coloneqq \{ \bm{R} \in \mathbb{R}^{3\times3} \mid \bm{R}\bm{R}^{\top} = \bm{R}^{\top}\bm{R} = \bm{I}, \det(\bm{R}) = 1\}
.
\end{align}
When the rotation matrix changes continuously over time $t$ around a fixed axis of rotation, the trajectory of the rotation matrix can be written as a function of time $\bm{R}(t)$, and the rotation matrix with a minute rotation is $\bm{R}(t+\Delta t) = \bm{R}(\Delta t)\bm{R}(t)$.
Therefore, the differential of the rotation matrix $\dot{\bm{R}}(t)$ is: 
\begin{align}
\dot{\bm{R}}(t) 
&= \lim_{\Delta t \to 0} \frac{\bm{R}(t+\Delta t) - \bm{R}(t)}{\Delta t}
\notag \\ &= 
\lim_{\Delta t \to 0} \frac{\bm{R}(\Delta t)\bm{R}(t) - \bm{R}(t)}{\Delta t}
\notag \\ &=
\lim_{\Delta t \to 0} \frac{\bm{R}(\Delta t) - \bm{I}}{\Delta t}\bm{R}(t)
\\
\bm{\Omega} &\coloneqq \lim_{\Delta t \to 0} \frac{\bm{R}(\Delta t) - \bm{I}}{\Delta t}
\\
\dot{\bm{R}}(t) &= \bm{\Omega}\bm{R}(t) \label{eq:dR}
.
\end{align}
The solution to equation \eqref{eq:dR} can be expressed as a matrix exponential function, as defined below:
\begin{align}
e^{\bm{\Omega}t} &\coloneqq \sum_{k=0}^{\infty} \frac{1}{k!}(\bm{\Omega}t)^k = \bm{R}(t).
\end{align}
The matrix logarithmic function is defined as the inverse of the matrix exponential function.
\begin{align}
\log(\bm{R}(t)) = \log(e^{\bm{\Omega}t}) \coloneqq \bm{\Omega}t \label{eq:log-def}
\end{align}

Because the rotation matrix is orthogonal,
\begin{align}
\bm{I} &= \bm{R}\bm{R}^{\top} = e^{\bm{\Omega}t}(e^{\bm{\Omega}t})^{\top} 
\notag \\ &=
(\bm{I} + \bm{\Omega}t + \cdots)(\bm{I} + \bm{\Omega}t + \cdots)^{\top}
\notag \\ &=
\bm{I} + (\bm{\Omega}+\bm{\Omega}^{\top})t + \cdots \\
\therefore \bm{\Omega}^{\top} &= -\bm{\Omega}.
\end{align}
This implies that $\bm{\Omega}$ is a skew-symmetric matrix.
That is, $\bm{\Omega}$ can be expressed as
\begin{align}
\bm{\Omega} &= 
\begin{bmatrix}
0 & -\omega_{z} & \omega_{y} \\
\omega_{z} & 0 & -\omega_{x} \\
-\omega_{y} & \omega_{x} & 0 \\
\end{bmatrix}
\notag \\ &= 
\omega_{x}
\begin{bmatrix}
0 & 0 & 0 \\
0 & 0 & -1 \\
0 & 1 & 0 \\
\end{bmatrix}
+
\omega_{y}
\begin{bmatrix}
0 & 0 & 1 \\
0 & 0 & 0 \\
-1 & 0 & 0 \\
\end{bmatrix}
+
\omega_{z}
\begin{bmatrix}
0 & -1 & 0 \\
1 & 0 & 0 \\
0 & 0 & 0 \\
\end{bmatrix}
.
\end{align}
Since a 3x3 skew-symmetric matrix can be expressed as a linear combination of three bases, it can also be described using the 3D vector $\bm{\omega} \in \mathbb{R}^3$.
The operators representing the conversion from the skew-symmetric matrix to the 3D vector $\vee$ and the conversion from the 3D vector to the skew-symmetric matrix $\wedge$ are defined as follows:
\begin{align}
\bm{\Omega}^{\vee}
&=
\begin{bmatrix}
0 & -\omega_{z} & \omega_{y} \\
\omega_{z} & 0 & -\omega_{x} \\
-\omega_{y} & \omega_{x} & 0 \\
\end{bmatrix}^{\vee}
\coloneqq
\begin{bmatrix}
\omega_{x} \\
\omega_{y} \\
\omega_{z} \\
\end{bmatrix}
=
\bm{\omega}
\\
\bm{\omega}^{\wedge}
&=
\begin{bmatrix}
\omega_{x} \\
\omega_{y} \\
\omega_{z} \\
\end{bmatrix}^{\wedge}
\coloneqq
\begin{bmatrix}
0 & -\omega_{z} & \omega_{y} \\
\omega_{z} & 0 & -\omega_{x} \\
-\omega_{y} & \omega_{x} & 0 \\
\end{bmatrix}
=
\bm{\Omega}.
\end{align}
This 3D vector $\bm{\omega}$ is called an ``angular velocity vector.''
If the time variable $t$ is fixed to one in the definition \eqref{eq:log-def}, $\bm{\omega}$ can be interpreted as a posture itself, not an angular velocity, and in this case, the 3D vector is called a ``rotation vector.''

At the end of this section, we present the relationship between the rotated angular velocity vector $\bm{R}\bm{\omega}$ and corresponding skew-symmetric matrix representation $\bm{R}\bm{\Omega}\bm{R}^{\top}$.
Using the correspondence between the skew-symmetric matrix and the cross-product ($\bm{\Omega}\bm{u}=\bm{\omega}\times\bm{u}, \bm{u}\in \mathbb{R}^3$),
\begin{align}
\bm{R}\bm{\Omega}\bm{R}^{\top} 
&=
\begin{bmatrix}
\bm{u_{x}}^{\top} \\
\bm{u_{y}}^{\top} \\
\bm{u_{z}}^{\top} \\
\end{bmatrix}
\bm{\Omega}
\begin{bmatrix}
\bm{u_{x}} & \bm{u_{y}} & \bm{u_{z}} 
\end{bmatrix}
\notag \\ &=
\begin{bmatrix}
\bm{u_{x}} \cdot (\bm{\omega} \times \bm{u_{x}}) & \bm{u_{x}} \cdot (\bm{\omega} \times \bm{u_{y}}) & \bm{u_{x}} \cdot (\bm{\omega} \times \bm{u_{z}}) \\
\bm{u_{y}} \cdot (\bm{\omega} \times \bm{u_{x}}) & \bm{u_{y}} \cdot (\bm{\omega} \times \bm{u_{y}}) & \bm{u_{y}} \cdot (\bm{\omega} \times \bm{u_{z}}) \\
\bm{u_{z}} \cdot (\bm{\omega} \times \bm{u_{x}}) & \bm{u_{z}} \cdot (\bm{\omega} \times \bm{u_{y}}) & \bm{u_{z}} \cdot (\bm{\omega} \times \bm{u_{z}}) \\
\end{bmatrix}
\notag \\ &=
\begin{bmatrix}
(\bm{u_{x}} \times \bm{u_{x}}) \cdot \bm{\omega} & (\bm{u_{y}} \times \bm{u_{x}}) \cdot \bm{\omega} & (\bm{u_{z}} \times \bm{u_{x}}) \cdot \bm{\omega} \\
(\bm{u_{x}} \times \bm{u_{y}}) \cdot \bm{\omega} & (\bm{u_{y}} \times \bm{u_{y}}) \cdot \bm{\omega} & (\bm{u_{z}} \times \bm{u_{y}}) \cdot \bm{\omega} \\
(\bm{u_{x}} \times \bm{u_{z}}) \cdot \bm{\omega} & (\bm{u_{y}} \times \bm{u_{z}}) \cdot \bm{\omega} & (\bm{u_{z}} \times \bm{u_{z}}) \cdot \bm{\omega} \\
\end{bmatrix}
\notag \\ &=
\begin{bmatrix}
0 & -\bm{u_{z}} \cdot \bm{\omega} & \bm{u_{y}} \cdot \bm{\omega} \\
\bm{u_{z}} \cdot \bm{\omega} & 0 & -\bm{u_{x}} \cdot \bm{\omega} \\
-\bm{u_{y}} \cdot \bm{\omega} & \bm{u_{x}} \cdot \bm{\omega} & 0 \\
\end{bmatrix}
\\
\therefore
(\bm{R}\bm{\Omega}\bm{R}^{\top})^{\vee}
&=
\begin{bmatrix}
\bm{u_{x}} \cdot \bm{\omega} \\
\bm{u_{y}} \cdot \bm{\omega} \\
\bm{u_{z}} \cdot \bm{\omega} \\
\end{bmatrix}
=
\bm{R}\bm{\omega}
\end{align}
where $\bm{u_{x}},\bm{u_{y}},\bm{u_{z}}$ are orthogonal unit vectors.
Because the rotation matrix is an orthogonal matrix, it can be expressed using three orthogonal unit vectors.


\subsection{Manipulator Dynamics}
In general, the manipulator dynamics can be expressed as
\begin{align}
\bm{\tau} ^ {ref} = \bm{M}(\bm{\theta})\ddot{\bm{\theta}} + \bm{h}(\bm{\theta}, \dot{\bm{\theta}}) + \bm{\tau} ^ {reac}
\end{align}
where 
$\bm{M}(\bm{\theta})$ is the inertial matrix,
$\bm{h}(\bm{\theta}, \dot{\bm{\theta}})$ includes the centrifugal, Coriolis, and gravitational forces, respectively, and
$\bm{\tau} ^ {ref}, \bm{\tau} ^ {reac}$ are the reference and reaction torques of each joint, respectively.
Currently, two manipulators must be used for teleoperation.
We can write the dynamics of the two manipulators using a single formula with a matrix as
\begin{align}
\begin{bmatrix}
\bm{\tau}_l \\
\bm{\tau}_f \\
\end{bmatrix} ^ {ref}
&= \begin{bmatrix}
\bm{M}_l & \bm{0} \\
\bm{0} & \bm{M}_f \\
\end{bmatrix}
\begin{bmatrix}
\ddot{\bm{\theta}}_l \\
\ddot{\bm{\theta}}_f \\
\end{bmatrix} 
+ 
\begin{bmatrix}
\bm{h}_l(\bm{\theta}_l, \dot{\bm{\theta}}_l) \\
\bm{h}_f(\bm{\theta}_f, \dot{\bm{\theta}}_f) \\
\end{bmatrix}
+
\begin{bmatrix}
\bm{\tau}_l \\
\bm{\tau}_f \\
\end{bmatrix} ^ {reac} 
\label{eq:manipulator-dynamics}
\end{align}
where the subscripts $l$ and $f$ represent leader and follower, respectively.
After compensating for $\bm{h}(\bm{\theta}, \dot{\bm{\theta}})$, 
\begin{align}
\begin{bmatrix}
\bm{\tau}_l \\
\bm{\tau}_f \\
\end{bmatrix} ^ {ref}
&=
\begin{bmatrix}
\bm{\tau}_l \\
\bm{\tau}_f \\
\end{bmatrix} ^ {ctrl}
+
\begin{bmatrix}
\hat{\bm{h}}_l(\bm{\theta}_l, \dot{\bm{\theta}}_l) \\
\hat{\bm{h}}_f(\bm{\theta}_f, \dot{\bm{\theta}}_f) \\
\end{bmatrix}
\label{eq:h-comp}
\end{align}
where $\hat{\bm{h}}(\bm{\theta}, \dot{\bm{\theta}})$ denotes the estimated value of $\bm{h}(\bm{\theta}, \dot{\bm{\theta}})$;
If the compensation is executed perfectly ($\bm{h}_l(\bm{\theta}_l, \dot{\bm{\theta}}_l) \simeq \hat{\bm{h}}_l(\bm{\theta}_l, \dot{\bm{\theta}}_l), \bm{h}_f(\bm{\theta}_f, \dot{\bm{\theta}}_f) \simeq \hat{\bm{h}}_f(\bm{\theta}_f, \dot{\bm{\theta}}_f)$), substituting equation \eqref{eq:h-comp} into equation \eqref{eq:manipulator-dynamics} gives
\begin{align}
\begin{bmatrix}
\bm{\tau}_l \\
\bm{\tau}_f \\
\end{bmatrix} ^ {ctrl}
&= \begin{bmatrix}
\bm{M}_l & \bm{0} \\
\bm{0} & \bm{M}_f \\
\end{bmatrix}
\begin{bmatrix}
\ddot{\bm{\theta}}_l \\
\ddot{\bm{\theta}}_f \\
\end{bmatrix} 
+ 
\begin{bmatrix}
\bm{\tau}_l \\
\bm{\tau}_f \\
\end{bmatrix} ^ {reac}.
\label{eq:two-manipulator-dynamics}
\end{align}
Even if there is an error in the compensation, it is included in the reaction force.

\section{Method}

\subsection{Control Target}

When scaling the leader's rotation with respect to the follower by a scaling factor $\alpha$, the goal of posture control is that the matrix obtained by multiplying the rotation matrix from the leader's base coordinate by the end-effector $\bm{R}_l$, and the inverse rotation matrix from the follower's base coordinate to the end effector by the scaling factor $\alpha$, $\bm{R}_f^{-\alpha}$, becomes an identity matrix.
In the rotation matrix representation, all rotations are expressed as rotation matrix multiplications; therefore, scaling is described as a power rather than a constant multiple.
Specifically,
\begin{align}
\bm{R}_l\bm{R}_f^{-\alpha} = \bm{I}.
\end{align}
Using a matrix logarithmic function, the posture error $\bm{\omega}_{e} \in \mathbb{R}^3$ can be expressed as 
\begin{align}
\bm{\omega}_{e} &\coloneqq 
\log 
\left (
    \bm{R}_l\bm{R}_f^{-\alpha}
\right )^{\vee}.
\end{align}
In addition, let $\bm{r}_l, \bm{r}_f \in \mathbb{R}^3$ be the position vector of the end effector of the leader and follower robots, and $\beta\coloneqq\rm{diag}(\beta_x, \beta_y, \beta_z)$ be the scaling factor for the translation; we define the error of the translation as follows:
\begin{align}
\bm{r}_{e} &\coloneqq \bm{r}_l - \beta \bm{r}_f
.
\end{align}
By integrating the posture and position errors, the position and posture control errors are defined as follows:
\begin{align}
\bm{x}_e &\coloneqq 
\begin{bmatrix}
\bm{\omega}_{e} &
\bm{r}_{e} \\
\end{bmatrix}^{\top}
.
\end{align}

In contrast, force control uses Newton's third law of motion; therefore, the goal of the control is for the sum of the wrenches on the leader and follower robots’ end effectors to be zero.
The wrench error $\bm{w}_e \in \mathbb{R}^6$ with scaling is shown below 
\begin{align}
\bm{w}_e &\coloneqq \bm{w}_{l} + \gamma \bm{w}_{f} \notag \\
&= \bm{J}_{l}^{+\top}\bm{\tau}_{l} + \gamma \bm{J}_{f}^{+\top}\bm{\tau}_{f} \notag \\
&= 
\begin{bmatrix}
\bm{J}_l^{+\top} & \gamma \bm{J}_f^{+\top} \\
\end{bmatrix}
\begin{bmatrix}
\bm{\tau}_l \\
\bm{\tau}_f \\
\end{bmatrix} ^ {reac}
\label{eq:wrench-error}
\end{align}
where $\bm{w}_{l},\bm{w}_{f}$ are the wrenches applied to the end-effector of each leader and follower, $\gamma=\rm{diag}(\gamma_{\tau x}, \gamma_{\tau y}, \gamma_{\tau z}, \gamma_{fx}, \gamma_{fy}, \gamma_{fz})$ is the scaling factor respectively. 
$\bm{J}_l,\bm{J}_f$ are the geometric Jacobian matrices of the leader and follower robots, defined by the relationship between the velocity $\dot{\bm{p}}$ and angular velocity $\bm{\omega}$ of the end-effector and the angular velocity of each joint $\dot{\bm{\theta}}$ as
$
\begin{bmatrix}
\bm{\omega} &
\bm{\dot{\bm{p}}} \\
\end{bmatrix}^{\top}
= \bm{J}\dot{\bm{\theta}}
$,
and $\bm{J}^{+}$ denotes the pseudoinverse matrix of $\bm{J}$.

\subsection{Error Dynamics}

The relationship between the dynamics of the manipulator and control objectives was investigated.
In general, the relationship between the position error vector $\bm{x}_e \in \mathbb{R}^6$ and the joint angles $\bm{\theta}_l, \bm{\theta}_f \in \mathbb{R}^6$ is a differentiable function $\rho : \mathbb{R}^6 \times \mathbb{R}^6\to\mathbb{R}^6$ such that $\bm{x}_e = \rho(\bm{\theta}_l, \bm{\theta}_f)$. Then, the time differential $\dot{\bm{x}}_e \in \mathbb{R}^6$ and second-order time differential $\ddot{\bm{x}}_e \in \mathbb{R}^6$ can be expressed by a block Jacobian matrix $
\begin{bmatrix}
\bm{J}_{x,l} & \bm{J}_{x,f}
\end{bmatrix}
=
\begin{bmatrix}
\partial \rho/\partial \theta_l & \partial \rho/\partial \theta_f
\end{bmatrix}
\in \mathbb{R}^{6\times12}
$ as follows:
\begin{align}
\dot{\bm{x}}_e &=
\begin{bmatrix}
\bm{J}_{x,l} & \bm{J}_{x,f}
\end{bmatrix}
\begin{bmatrix}
\dot{\bm{\theta}}_l \\
\dot{\bm{\theta}}_f \\
\end{bmatrix} 
\\
\ddot{\bm{x}}_e &=
\begin{bmatrix}
\bm{J}_{x,l} & \bm{J}_{x,f}
\end{bmatrix}
\begin{bmatrix}
\ddot{\bm{\theta}}_l \\
\ddot{\bm{\theta}}_f \\
\end{bmatrix} 
+
\begin{bmatrix}
\dot{\bm{J}}_{x,l} & \dot{\bm{J}}_{x,f}
\end{bmatrix}
\begin{bmatrix}
\dot{\bm{\theta}}_l \\
\dot{\bm{\theta}}_f \\
\end{bmatrix}.
\label{eq:ddot-x_e}
\end{align}
Substituting the dynamics equation of the manipulator \eqref{eq:two-manipulator-dynamics} into the angular acceleration vector of \eqref{eq:ddot-x_e} yields
\begin{align}
\ddot{\bm{x}}_e
&=
\begin{bmatrix}
\bm{J}_{x,l}\bm{M}_l ^ {-1} & \bm{J}_{x,f}\bm{M}_f ^ {-1}
\end{bmatrix}
\left \{
\begin{bmatrix}
\bm{\tau}_l \\
\bm{\tau}_f \\
\end{bmatrix} ^ {ctrl}
- 
\begin{bmatrix}
\bm{\tau}_l \\
\bm{\tau}_f \\
\end{bmatrix} ^ {reac}
\right \}
\notag \\ & \qquad +
\begin{bmatrix}
\dot{\bm{J}}_{x,l} & \dot{\bm{J}}_{x,f}
\end{bmatrix}
\begin{bmatrix}
\dot{\bm{\theta}}_l \\
\dot{\bm{\theta}}_f \\
\end{bmatrix}.
\label{eq:ddot-pos-error-dynamics}
\end{align}

\begin{figure*}[t]
    \centering
    \includegraphics[width=\linewidth]{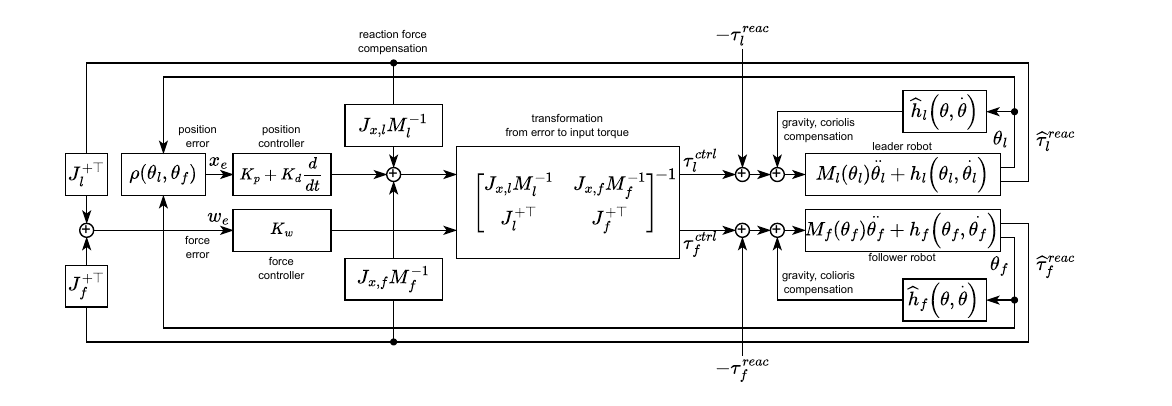}
    \caption{Block diagram of 4-channel bilateral controller}
    \label{fig:controller_block_diagram}
\end{figure*}

The rotation matrix $\bm{R}$ of the end-effector posture can be expressed as
a function of the angle of each joint $\bm{R}(\bm{\theta})$,
so the posture error function $\rho_r : \mathbb{R}^6 \times \mathbb{R}^6 \to \mathbb{R}^3$ is $\bm{\omega}_e = \rho_r(\bm{\theta}_l, \bm{\theta}_f) =\log(\bm{R}_l(\bm{\theta}_l)\bm{R}_l^{-\alpha}(\bm{\theta}_l))^{\vee}$.
The differential of the posture error function is expressed as follows:
\begin{align}
\dot{\bm{\omega}}_{e}^{\wedge} &=
\frac{d}{dt}
\log 
    \left (
        \bm{R}_l\bm{R}_f^{-\alpha}
    \right )
\notag \\
&= 
\left (
    \bm{R}_l\bm{R}_f^{-\alpha}
\right )^{-1}
\left (
    \dot{\bm{R}_l}\bm{R}_f^{-\alpha} + \bm{R}_l\dot{\bm{R}}_f^{-\alpha}
\right )
\notag \\
&=
\bm{R}_f^{\alpha}\bm{R}_l^{-1}
\left (
    \dot{\bm{R}_l} \bm{R}_f^{-\alpha} - \bm{R}_l\sum_{i=0}^{\alpha-1}\bm{R}_f^{-i-1} \dot{\bm{R}}_f \bm{R}_f^{-\alpha+i}
\right )
\notag \\
&= 
\bm{R}_f^{\alpha}\bm{R}_l^{-1}
    \dot{\bm{R}_l} \bm{R}_f^{-\alpha} - \sum_{i=0}^{\alpha-1}\bm{R}_f^{-i-1+\alpha} \dot{\bm{R}}_f \bm{R}_f^{-\alpha+i}
\notag \\
&= 
\bm{R}_f^{\alpha}\bm{R}_l^{-1}\bm{\Omega}_l\bm{R}_l\bm{R}_f^{-\alpha} - \sum_{i=0}^{\alpha-1}\bm{R}_f^{(\alpha-1-i)} \bm{\Omega}_f \bm{R}_f^{-(\alpha-1-i)}
\notag \\
&= 
\bm{R}_f^{\alpha}\bm{R}_l^{-1}\bm{\Omega}_l\bm{R}_l\bm{R}_f^{-\alpha} - \sum_{i=0}^{\alpha-1}\bm{R}_f^{i} \bm{\Omega}_f \bm{R}_f^{-i}
.
\end{align}
When rewritten in terms of three-dimensional vectors, we obtain
\begin{align}
\dot{\bm{\omega}}_{e} &= 
(\bm{R}_f^{\alpha}\bm{R}_l^{-1}\bm{\Omega}_l\bm{R}_l\bm{R}_f^{-\alpha})^{\vee} - \sum_{i=0}^{\alpha-1} (\bm{R}_f^{i} \bm{\Omega}_f \bm{R}_f^{-i})^{\vee}
\notag \\ &=
\bm{R}_f^{\alpha}\bm{R}_l^{-1}\bm{\omega}_l - \sum_{i=0}^{\alpha-1}\bm{R}_f^{i} \bm{\omega}_f
.
\end{align}
The differential of the transformation error is 
\begin{align}
\dot{\bm{p}}_{e} &= \dot{\bm{p}}_l - \beta \dot{\bm{p}}_f 
.
\end{align}
Therefore, the Jacobian matrices of the position and posture control error functions are as follows.
\begin{align}
\dot{\bm{x}}_{e} &= 
\begin{bmatrix}
\dot{\bm{\omega}}_{e} \\
\dot{\bm{p}}_{e} \\
\end{bmatrix}
= 
\begin{bmatrix}
\bm{R}_f^{\alpha}\bm{R}_l^{-1}\bm{\omega}_l - \sum_{i=0}^{\alpha-1}\bm{R}_f^{i}\bm{\omega}_f \\
\dot{\bm{p}}_l - \beta \dot{\bm{p}}_f \\
\end{bmatrix}
\notag \\ &= 
\begin{bmatrix}
\bm{R}_f^{\alpha}\bm{R}_l^{-1} & \bm{0} \\
\bm{0} & \bm{I}
\end{bmatrix}
\begin{bmatrix}
\bm{\omega}_l \\
\dot{\bm{p}}_l \\
\end{bmatrix}
-
\begin{bmatrix}
\sum_{i=0}^{\alpha-1}\bm{R}_f^{i} & \bm{0} \\
\bm{0} & \beta
\end{bmatrix}
\begin{bmatrix}
\bm{\omega}_f \\
\dot{\bm{p}}_f \\
\end{bmatrix}
\notag \\ &= 
\begin{bmatrix}
\bm{R}_f^{\alpha}\bm{R}_l^{-1} & \bm{0} \\
\bm{0} & \bm{I}
\end{bmatrix}
\bm{J}_{l}\dot{\bm{\theta}}_l
-
\begin{bmatrix}
\sum_{i=0}^{\alpha-1}\bm{R}_f^{i} & \bm{0} \\
\bm{0} & \beta
\end{bmatrix}
\bm{J}_{f}\dot{\bm{\theta}}_f
\notag \\ &= 
\begin{bmatrix}
\begin{bmatrix}
\bm{R}_f^{\alpha}\bm{R}_l^{-1} & \bm{0} \\
\bm{0} & \bm{I}
\end{bmatrix}
\bm{J}_{l}
& - 
\begin{bmatrix}
\sum_{i=0}^{\alpha-1}\bm{R}_f^{i} & \bm{0} \\
\bm{0} & \beta
\end{bmatrix}
\bm{J}_{f}
\end{bmatrix}
\begin{bmatrix}
\dot{\bm{\theta}}_l \\
\dot{\bm{\theta}}_f \\
\end{bmatrix}
\end{align}
\begin{align}
\bm{J}_{x,l} &= 
\begin{bmatrix}
\bm{R}_f^{\alpha}\bm{R}_l^{-1} & \bm{0} \\
\bm{0} & \bm{I}
\end{bmatrix}
\bm{J}_{l} 
, \quad
\bm{J}_{x,f}
= 
- 
\begin{bmatrix}
\sum_{i=0}^{\alpha-1}\bm{R}_f^{i} & \bm{0} \\
\bm{0} & \beta
\end{bmatrix}
\bm{J}_{f}
.
\end{align}

Substituting the manipulator's dynamics equation \eqref{eq:two-manipulator-dynamics} into the wrench error function \eqref{eq:wrench-error}, we obtain
\begin{align}
\bm{w}_e &= 
\begin{bmatrix}
\bm{J}_l^{+\top} & \gamma \bm{J}_f^{+\top} \\
\end{bmatrix}
\begin{bmatrix}
\bm{\tau}_l \\
\bm{\tau}_f \\
\end{bmatrix} ^ {reac}
\notag \\ &=
\begin{bmatrix}
\bm{J}_l^{+\top} & \gamma \bm{J}_f^{+\top} \\
\end{bmatrix}
\left \{
\begin{bmatrix}
\bm{\tau}_l \\
\bm{\tau}_f \\
\end{bmatrix} ^ {ctrl}
-
\begin{bmatrix}
\bm{M}_l & \bm{0} \\
\bm{0} & \bm{M}_f \\
\end{bmatrix}
\begin{bmatrix}
\ddot{\bm{\theta}}_l \\
\ddot{\bm{\theta}}_f \\
\end{bmatrix}
\right \}.
\label{eq:wrench-error-dynamics}
\end{align}

The dynamic equations for the position \eqref{eq:ddot-pos-error-dynamics} and wrench errors \eqref{eq:wrench-error-dynamics} can be written together as a matrix as follows:
\begin{align}
\begin{bmatrix}
\ddot{\bm{x}}_e \\
\bm{w}_e  \\
\end{bmatrix} &=
\begin{bmatrix}
\bm{J}_{x,l}\bm{M}_l^{-1} & \bm{J}_{x,f}\bm{M}_f^{-1} \\
\bm{J}_l^{+\top} & \gamma \bm{J}_f^{+\top} \\
\end{bmatrix}
\begin{bmatrix}
\bm{\tau}_l \\
\bm{\tau}_f \\
\end{bmatrix} ^ {ctrl}
\notag \\ & \qquad - 
\begin{bmatrix}
\bm{J}_{x,l}\bm{M}_l^{-1} & \bm{J}_{x,f}\bm{M}_f^{-1} \\
\bm{0} & \bm{0} \\
\end{bmatrix}
\begin{bmatrix}
\bm{\tau}_l \\
\bm{\tau}_f \\
\end{bmatrix} ^ {reac}
\notag \\ & \qquad +
\begin{bmatrix}
\dot{\bm{J}}_{x,l} & \dot{\bm{J}}_{x,f} \\
\bm{0} & \bm{0} \\
\end{bmatrix}
\begin{bmatrix}
\dot{\bm{\theta}}_l \\
\dot{\bm{\theta}}_f \\
\end{bmatrix}
-
\begin{bmatrix}
\bm{0} & \bm{0} \\
\bm{J}_l^{+\top} & \gamma \bm{J}_f^{+\top} \\
\end{bmatrix}
\begin{bmatrix}
\bm{M}_l & \bm{0} \\
\bm{0} & \bm{M}_f \\
\end{bmatrix}
\begin{bmatrix}
\ddot{\bm{\theta}}_l \\
\ddot{\bm{\theta}}_f \\
\end{bmatrix}
\\
\bm{H} &\coloneqq 
\begin{bmatrix}
\bm{J}_{x,l}\bm{M}_l^{-1} & \bm{J}_{x,f}\bm{M}_f^{-1} \\
\bm{J}_l^{+\top} & \gamma \bm{J}_f^{+\top} \\
\end{bmatrix}
\\
\begin{bmatrix}
\ddot{\bm{x}}_e \\
\bm{w}_e  \\
\end{bmatrix} 
&=
\bm{H}
\begin{bmatrix}
\bm{\tau}_l \\
\bm{\tau}_f \\
\end{bmatrix} ^ {ctrl}
- 
\begin{bmatrix}
\bm{I} & \bm{0} \\
\bm{0} & \bm{0} \\
\end{bmatrix}
\bm{H}
\begin{bmatrix}
\bm{\tau}_l \\
\bm{\tau}_f \\
\end{bmatrix} ^ {reac}
+
\begin{bmatrix}
\dot{\bm{J}}_{x,l} & \dot{\bm{J}}_{x,f} \\
\bm{0} & \bm{0} \\
\end{bmatrix}
\begin{bmatrix}
\dot{\bm{\theta}}_l \\
\dot{\bm{\theta}}_f \\
\end{bmatrix}
\notag \\ & \qquad -
\begin{bmatrix}
\bm{0} & \bm{0} \\
\bm{0} & \bm{I} \\
\end{bmatrix}
\bm{H}
\begin{bmatrix}
\bm{M}_l & \bm{0} \\
\bm{0} & \bm{M}_f \\
\end{bmatrix}
\begin{bmatrix}
\ddot{\bm{\theta}}_l \\
\ddot{\bm{\theta}}_f \\
\end{bmatrix}.
\label{eq:total-error-dynamics}
\end{align}

\subsection{Controller Design}
A block diagram of the proposed controller is shown in Fig~\ref{fig:controller_block_diagram}.
If we transform the error dynamics equation \eqref{eq:total-error-dynamics} and place the vector of joint torques applied by the actuator on the left-hand side, we obtain
\begin{align}
\begin{bmatrix}
\bm{\tau}_l \\
\bm{\tau}_f \\
\end{bmatrix} ^ {ctrl}
&=
\bm{H} ^ {-1}
\begin{bmatrix}
\ddot{\bm{x}}_e \\
\bm{w}_e \\
\end{bmatrix}
+ 
\bm{H} ^ {-1}
\begin{bmatrix}
\bm{I} & \bm{0} \\
\bm{0} & \bm{0} \\
\end{bmatrix}
\bm{H}
\begin{bmatrix}
\bm{\tau}_l \\
\bm{\tau}_f \\
\end{bmatrix} ^ {reac}
\notag \\ & \qquad -
\bm{H} ^ {-1}
\begin{bmatrix}
\dot{\bm{J}}_{x,l} & \dot{\bm{J}}_{x,f} \\
\bm{0} & \bm{0} \\
\end{bmatrix}
\begin{bmatrix}
\dot{\bm{\theta}}_l \\
\dot{\bm{\theta}}_f \\
\end{bmatrix}
\notag \\ & \qquad +
\bm{H} ^ {-1}
\begin{bmatrix}
\bm{0} & \bm{0} \\
\bm{0} & \bm{I} \\
\end{bmatrix}
\bm{H}
\begin{bmatrix}
\bm{M}_l & \bm{0} \\
\bm{0} & \bm{M}_f \\
\end{bmatrix}
\begin{bmatrix}
\ddot{\bm{\theta}}_l \\
\ddot{\bm{\theta}}_f \\
\end{bmatrix}
.
\end{align}
Because the values of terms 3 and 4 decrease at low speeds, ($\dot{\bm{\theta}}_l\simeq0,\dot{\bm{\theta}}_f\simeq0,\ddot{\bm{\theta}}_l\simeq0,\ddot{\bm{\theta}}_f\simeq0$), they are omitted and substituted into equation \eqref{eq:h-comp},
\begin{align}
\begin{bmatrix}
\bm{\tau}_l \\
\bm{\tau}_f \\
\end{bmatrix} ^ {ref}
&=
\bm{H} ^ {-1}
\begin{bmatrix}
\ddot{\bm{x}}_e \\
\bm{w}_e \\
\end{bmatrix}
+ 
\bm{H} ^ {-1}
\begin{bmatrix}
\bm{I} & \bm{0} \\
\bm{0} & \bm{0} \\
\end{bmatrix}
\bm{H}
\begin{bmatrix}
\bm{\tau}_l \\
\bm{\tau}_f \\
\end{bmatrix} ^ {reac}
+
\begin{bmatrix}
\hat{\bm{h}}_l(\bm{\theta}_l, \dot{\bm{\theta}}_l) \\
\hat{\bm{h}}_f(\bm{\theta}_f, \dot{\bm{\theta}}_f) \\
\end{bmatrix}.
\label{eq:controller1}
\end{align}

By substituting the desired error dynamics equation into $\ddot{\bm{x}}_e$ and $\bm{w}_e$ in \eqref{eq:controller1}, the control input is obtained. For example, if we use proportional derivative control for position control $\ddot{\bm{x}}_e^{ref} = -K_p \bm{x}_e - K_d \dot{\bm{x}}_e$,
and proportional (P) control for force control $\bm{w}_e^{ref} = - K_{w}\bm{w}_e$,
The controller is obtained as follows:
\begin{align}
\begin{bmatrix}
\bm{\tau}_l \\
\bm{\tau}_f \\
\end{bmatrix} ^ {ref}
&=
\bm{H} ^ {-1}
\begin{bmatrix}
-K_{p} \bm{x}_e - K_{d} \dot{\bm{x}}_e \\
- K_{w}\bm{w}_e \\
\end{bmatrix}
\notag \\ & \qquad + 
\bm{H} ^ {-1}
\begin{bmatrix}
\bm{I} & \bm{0} \\
\bm{0} & \bm{0} \\
\end{bmatrix}
\bm{H}
\begin{bmatrix}
\bm{\tau}_l \\
\bm{\tau}_f \\
\end{bmatrix} ^ {reac}
+
\begin{bmatrix}
\hat{\bm{h}}_l(\bm{\theta}_l, \dot{\bm{\theta}}_l) \\
\hat{\bm{h}}_f(\bm{\theta}_f, \dot{\bm{\theta}}_f) \\
\end{bmatrix}
\end{align}

Here, $K_{p}$ and $K_{d}$ represent the parameters of the virtual spring and damper, respectively, between the end effectors of the leader and follower robots.
$K_{w}$ denotes the virtual mass felt by the operator.
If $K_{w}=0$, the operator perceives the sum of the masses of the leader and follower robots.
Although higher values of $K_{w}$ result in a lighter feeling for the operator, the effects of the force observation noise increase.

\begin{figure*}[t]
    \centering
    \begin{minipage}[t]{0.49\linewidth}
        \centering
        \includegraphics[width=\linewidth]{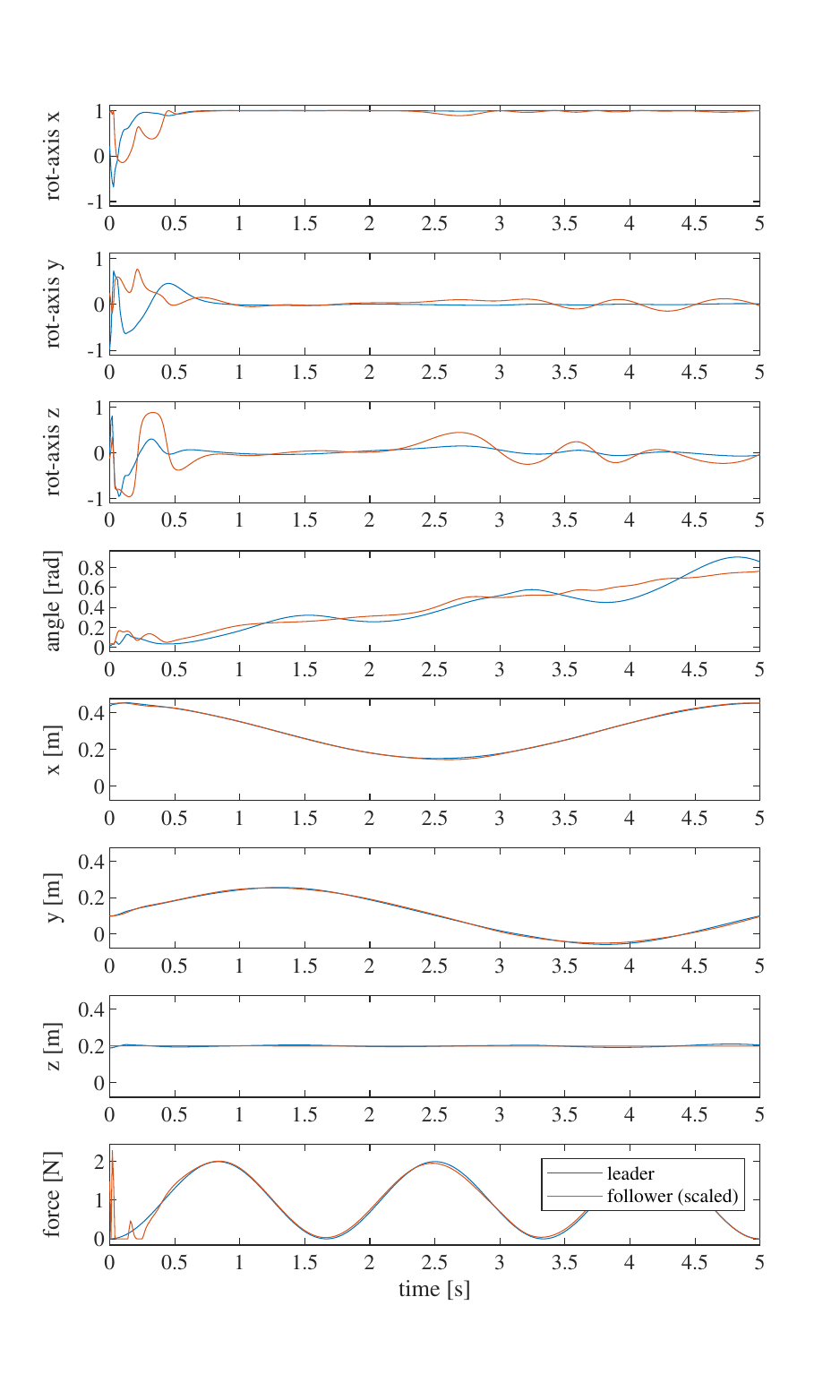}
        \subcaption{Rotation vector error}
    \end{minipage}
    \begin{minipage}[t]{0.49\linewidth}
        \centering
        \includegraphics[width=\linewidth]{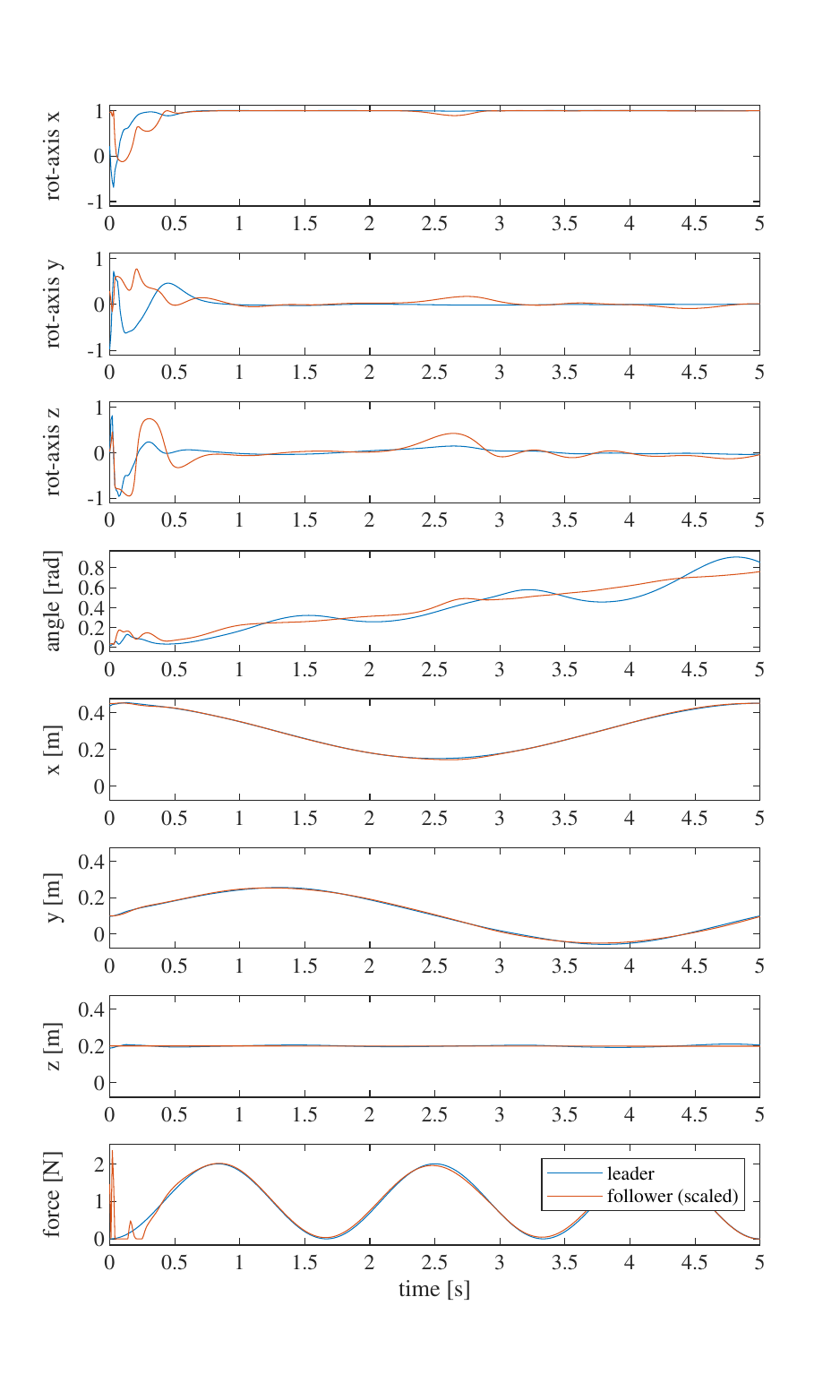}
        \subcaption{Rotation matrix error}
    \end{minipage}
    \caption{Simulation results}
    \label{fig:simulation_results}
\end{figure*}

\section{Numerical Example}

\subsection{Simulation setup}
We conducted simulation experiments using MATLAB~\cite{MATLAB} and the Robotics System Toolbox~\cite{RoboticsSystemsToolbox}.
We used a 7-DOF CRANE-X7 manipulator as the 6-DOF manipulator by fixing the third joint to the base link.
We simulated the reaction force of the leader robot using a position/force hybrid controller that corresponded to human operation in the real world.
The target trajectory was set as a circular path in the x-y plane with the end effector rotating, and a target in the z-direction was given as a force.

The desired trajectory was as follows:
\begin{align}
\bm{p}^{d}
&=
\begin{bmatrix}
x \\
y \\
z \\
\end{bmatrix} ^ {d}
=
\begin{bmatrix}
r \cos(\frac{2\pi}{T} t) \\
r \sin(\frac{2\pi}{T} t) \\
h \beta \\
\end{bmatrix}
\end{align}
where t is time, 
and the desired velocity was
$\bm{v}^{d} = \bm{p}^{d} - \bm{p}_l$.
The Euler angle representation (XYZ) generated the desired rotation as follows:
\begin{align}
\begin{bmatrix}
\theta_x \\
\theta_y \\
\theta_y \\
\end{bmatrix} ^ {d}
=
\begin{bmatrix}
\frac{\pi}{4} \cdot \frac{t}{T} \\
0 \\
0 \\
\end{bmatrix}.
\end{align}
Here, the desired angular velocity vector can be calculated as
\begin{align}
\bm{\omega}^{d} = \bm{R}_l (\log (\bm{R}_l^{\top}\bm{R}^{ref}))^{\vee}.
\end{align}
If the P gain for trajectory tracking is $K_p^{traj}$, the desired angular velocity of each joint $\dot{\bm{\theta}}^{d}$ can be calculated using the geometric Jacobian matrix as follows: 
\begin{align}
\dot{\bm{\theta}}^{d} = K_p^{traj} \bm{J}_l^{+}
\begin{bmatrix}
\bm{\omega}^{ref} \\
\bm{v}^{ref} \\
\end{bmatrix}.
\end{align}
If the D gain for trajectory tracking is $K_d^{traj}$, the torque of each joint for trajectory tracking $\bm{\tau}^{traj}$ is
\begin{align}
\bm{\tau}^{traj} = K_d^{traj} \bm{M}_l (\dot{\bm{\theta}}^{ref} - \dot{\bm{\theta}}_{l}).
\end{align}
Finally, the reaction force of the leader robot $\bm{\tau}^{reac}_l$ is 
\begin{align}
\bm{\tau}^{reac}_l &= \bm{J}_l \{ \bm{S} \bm{J}_l^{+} \bm{\tau}^{traj} + (\bm{I} - \bm{S}) \bm{w}^d \}
\end{align}
where $\bm{S} = \mathrm{diag}(1,1,1,1,1,0)$ is the selection matrix and $\bm{w}^d = \mathrm{diag}(0,0,0,0,0,1 - \cos(\frac{6\pi}{T} t))$ is the desired force.

To simulate the follower robot, we simulated robot dynamics with a constrained surface (in this case, the offset z-plane) as follows:
\begin{align}
\bm{\tau}^{reac}_f &= 
\begin{cases}
0 & (z_{f} > h) \\
\bm{J}_l(\bm{I} - \bm{S})\bm{J}_l^{+} \bm{\tau}_{l}^{ctrl} & (z_{f} \leq h)
\end{cases}
\\
\begin{bmatrix}
\bm{\omega}_f \\
\bm{v}_f \\
\end{bmatrix}
&= 
\begin{cases}
\bm{J}_l \dot{\bm{\theta_{l}}} & (z_{f} > h) \\
\bm{S}\bm{J}_l \dot{\bm{\theta_{l}}} & (z_{f} \leq h)
\end{cases}
.
\end{align}

\begin{figure}[t]
    \centering
    \includegraphics[width=\linewidth]{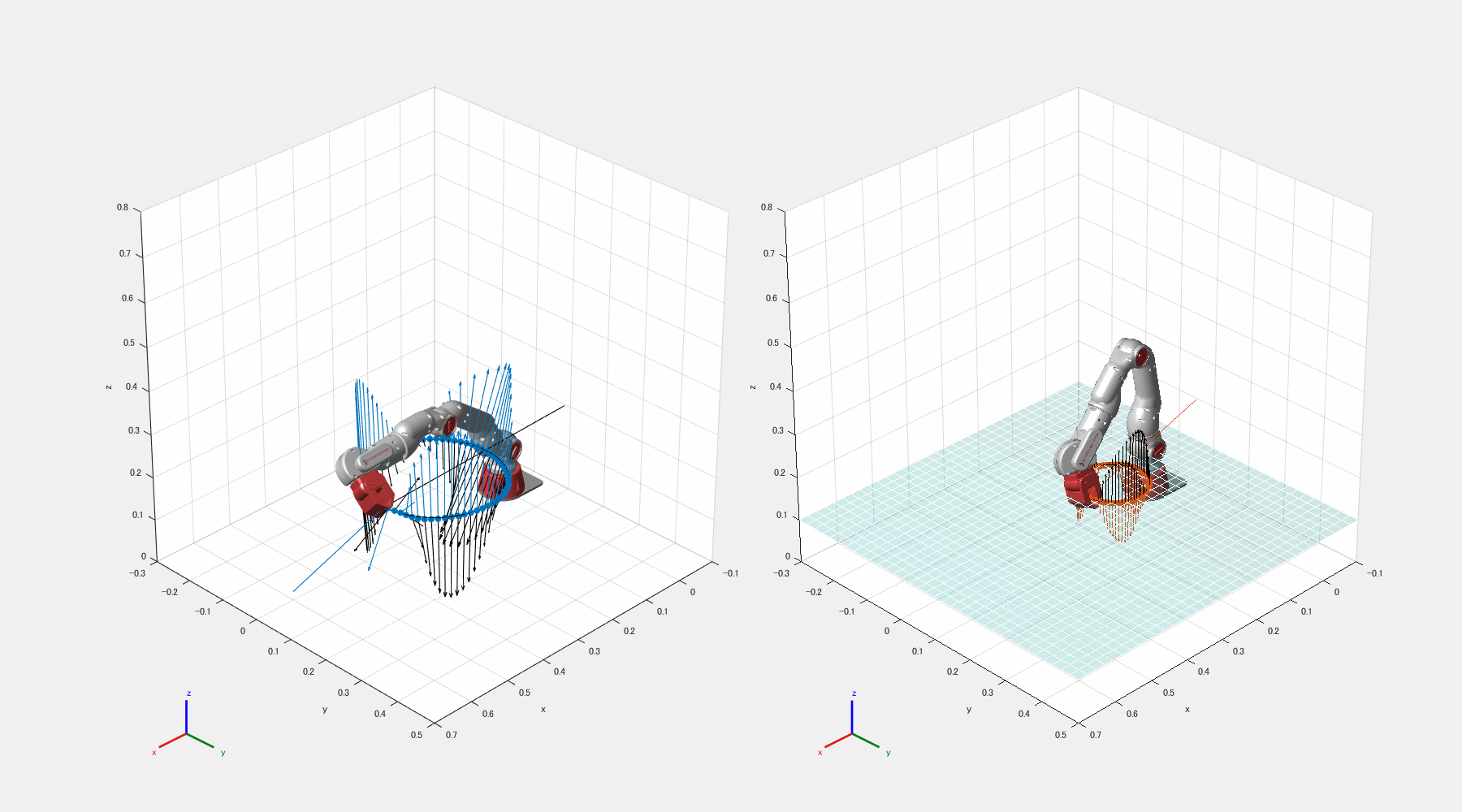}
    \caption{Trajectory overview}
    \label{fig:trajectory_overview}
\end{figure}

\begin{figure}[t]
    \centering
    \includegraphics[width=\linewidth]{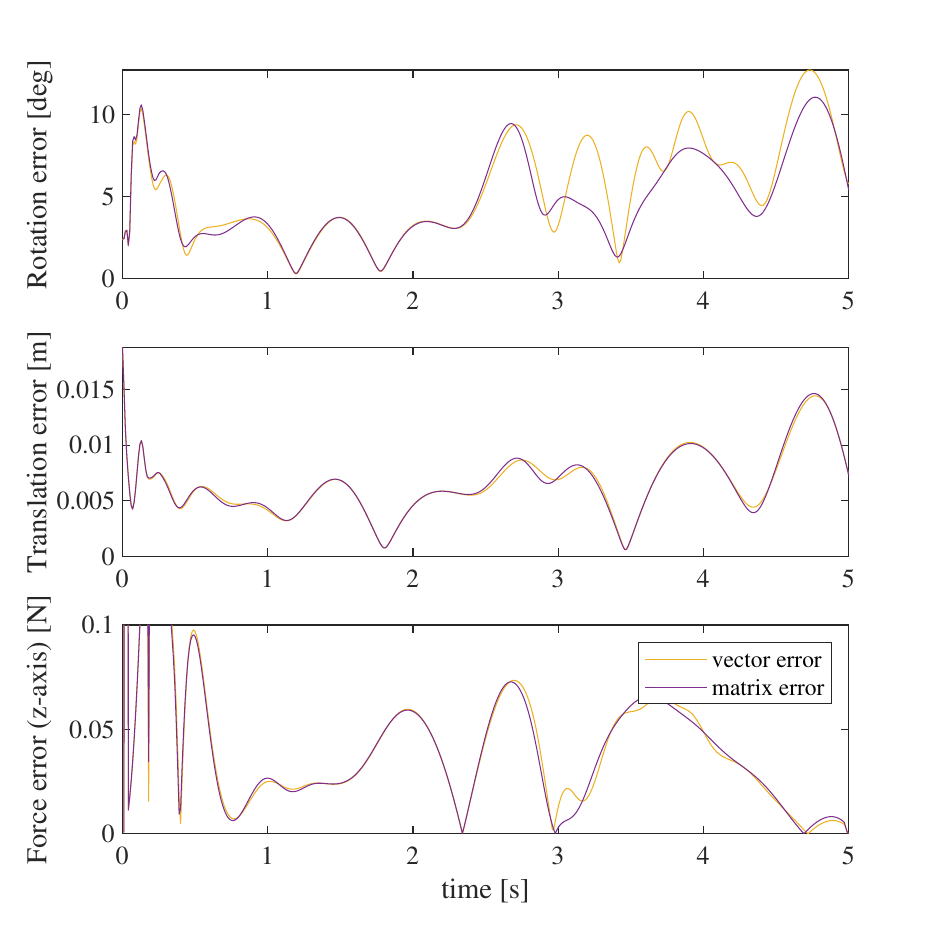}
    \caption{Error norm}
    \label{fig:error_norm}
\end{figure}

\vspace{2mm}
\subsection{Simulation results}

The simulation results are presented in Fig.~\ref{fig:simulation_results}.
The results for the follower robot were scaled on a graph using a specified scaling factor.
In addition, the trajectory overview is shown in Fig.~\ref{fig:trajectory_overview}; the norm of rotation, translation, and wrench error is shown in Fig~\ref{fig:error_norm}, and the parameters are listed in Table~\ref{table:simulation_parameters}.
We confirmed that the position and force followed a specified scaling factor.
We compared two posture errors: the rotation vector error, which is a simple difference in the rotation vector ($\bm{\omega}_e = \log(\bm{R}_l)^{\vee} - \alpha \log(\bm{R}_f)^{\vee}$), and the rotation matrix error as the proposed method ($
\bm{\omega}_e = 
\log 
\left (
    \bm{R}_l\bm{R}_f^{-\alpha}
\right )
^{\vee}
$).
For the vector error, the Jacobian matrix of the position and posture control error functions is as follows:
\begin{align}
\bm{J}_{x,l} &= 
\bm{J}_{l} 
, \quad
\bm{J}_{x,f}
= 
- 
\begin{bmatrix}
\alpha \bm{I} & \bm{0} \\
\bm{0} & \beta
\end{bmatrix}
\bm{J}_{f}
.
\end{align}

Overall, translation showed no significant difference; however, the rotation control performance was different, and the matrix error was better than the vector error.
Specifically, in Fig.~\ref{fig:simulation_results}, the amplitude of the vibration of the y- and z-axis of rotations in the matrix error case was small.
In addition, in Fig.~\ref{fig:error_norm}, the rotation error of the matrix error case is typically smaller than that of the vector error case.
This difference occurs because the matrix error uses the shortest path between two rotation matrices, whereas the vector error does not use the exact shortest path.

\section{Conclusion}
This study proposes a scaling 4-channel bilateral control on the Cartesian coordinate system that decouples each dimension in the Cartesian coordinate system regardless of the scaling factor and achieves the desired error dynamics.
Future issues include determining the robustness to observation noise and modeling errors, determining the effect of the control period on the actual implementation with a sampled value control system, handling redundant degrees of freedom, and verification by simulation and actual devices.

\begin{table}[t]
    \caption{Simulation Parameters}
    \label{table:simulation_parameters}
    \centering
    \begin{tabular}{clc}
        \hline
         & Parameter & Value \\
        \hline
        $K_{p}$ & P gain for position control & 100 \\
        $K_{d}$ & D gain for position control & 10 \\
        $K_{w}$ & P gain for force control & 0.1 \\
        $K_{p}^{traj}$ & P gain for trajectory tracking & 50 \\
        $K_{d}^{traj}$ & D gain for trajectory tracking & 50 \\
        $\alpha$ & Scaling factor for rotation & 2 \\
        $\beta$ & Scaling factor for translation & 2 \\
        $\gamma$ & Scaling factor for wrench & 2 \\
        $f$ & Sampling frequency [Hz] & 100 \\
        $T$ & Length of time to simulate [s] & 5 \\
        $r$ & Radius of reference trajectory [m] & 0.15 \\
        $h$ & Height of constraint surface [m] & 0.1 \\
        \hline
    \end{tabular}
\end{table}

\acknowledgment
This work was supported by JSPS KAKENHI Grant Num-
ber 24K00905, JST, PRESTO Grant Number JPMJPR24T3
Japan and JST ALCA-Next Japan, Grant Number JPM-
JAN24F1. This study was based on the results obtained from
the JPNP20004 project subsidized by the New Energy and
Industrial Technology Development Organization (NEDO).


\bibliographystyle{IEEEtran}
\bibliography{references}

\begin{thebibliography}{1}
\providecommand{\url}[1]{#1}
\csname url@samestyle\endcsname
\providecommand{\newblock}{\relax}
\providecommand{\bibinfo}[2]{#2}
\providecommand{\BIBentrySTDinterwordspacing}{\spaceskip=0pt\relax}
\providecommand{\BIBentryALTinterwordstretchfactor}{4}
\providecommand{\BIBentryALTinterwordspacing}{\spaceskip=\fontdimen2\font plus
\BIBentryALTinterwordstretchfactor\fontdimen3\font minus \fontdimen4\font\relax}
\providecommand{\BIBforeignlanguage}[2]{{%
\expandafter\ifx\csname l@#1\endcsname\relax
\typeout{** WARNING: IEEEtran.bst: No hyphenation pattern has been}%
\typeout{** loaded for the language `#1'. Using the pattern for}%
\typeout{** the default language instead.}%
\else
\language=\csname l@#1\endcsname
\fi
#2}}
\providecommand{\BIBdecl}{\relax}
\BIBdecl

\bibitem{tsuji2006controller}
T.~Tsuji, K.~Natori, H.~Nishi, and K.~Ohnishi, ``A controller design method of bilateral control system,'' \emph{EPE Journal}, vol.~16, no.~2, pp. 22--28, 2006.

\bibitem{sakaino2011multi}
S.~Sakaino, T.~Sato, and K.~Ohnishi, ``Multi-dof micro-macro bilateral controller using oblique coordinate control,'' \emph{IEEE Transactions on Industrial Informatics}, vol.~7, no.~3, pp. 446--454, 2011.

\bibitem{suzuki2018development}
H.~Suzuki, H.~Masuda, K.~Hongo, R.~Horie, S.~Yajima, Y.~Itotani, M.~Fujita, and K.~Nagasaka, ``Development and testing of force-sensing forceps using fbg for bilateral micro-operation system,'' \emph{IEEE Robotics and Automation Letters}, vol.~3, no.~4, pp. 4281--4288, 2018.

\bibitem{yilmaz2023sensorless}
N.~Yilmaz, B.~Burkhart, A.~Deguet, P.~Kazanzides, and U.~Tumerdem, ``Sensorless transparency optimized haptic teleoperation on the da vinci research kit,'' \emph{IEEE Robotics and Automation Letters}, vol.~9, no.~2, pp. 971--978, 2024.

\bibitem{michel2024passivity}
Y.~Michel, Y.~Abdelhalem, and G.~Cheng, ``Passivity-based teleoperation with variable rotational impedance control,'' \emph{IEEE Robotics and Automation Letters}, pp. 1--8, 2024.

\bibitem{MATLAB}
\BIBentryALTinterwordspacing
T.~M. Inc., ``Matlab version: 24.1.0 (r2024a),'' Natick, Massachusetts, United States, 2024. [Online]. Available: \url{https://www.mathworks.com}
\BIBentrySTDinterwordspacing

\bibitem{RoboticsSystemsToolbox}
\BIBentryALTinterwordspacing
------, ``Robotics system toolbox version: 24.1 (r2024a),'' Natick, Massachusetts, United States, 2024. [Online]. Available: \url{https://www.mathworks.com}
\BIBentrySTDinterwordspacing

\end{thebibliography}


\end{document}